\documentclass[5p]{elsarticle}

\usepackage{lineno,hyperref}
\modulolinenumbers[5]

\journal{Journal of \LaTeX\ Templates}

\usepackage{amssymb}
\usepackage{amsmath}
\usepackage{float}
\usepackage{subfig}
\usepackage{color}
\usepackage{multirow}

\begin{document}

\begin{frontmatter}

\title{Effective Actor-centric Human-object Interaction Detection}

\author{Kunlun Xu}
\author{Zhimin Li}
\author{Zhijun Zhang}
\author{Leizhen Dong}

\author{Wenhui Xu}
\author{Luxin Yan}
\author{Sheng Zhong}

\author{Xu Zou\corref{mycorrespondingauthor}}

\address{School of Artificial Intelligence and Automation, Huazhong University of Science and Technology, Wuhan 430074, China}
%\cortext[mycorrespondingauthor]{Corresponding author}
%\ead{zoux@hust.edu.cn}

\begin{abstract}
While Human-Object Interaction(HOI) Detection has achieved tremendous advances in recent, it still remains challenging due to complex interactions with multiple humans and objects occurring in images, which would inevitably lead to ambiguities. 
Most existing methods either generate all human-object pair candidates and infer their relationships by cropped local features successively in a two-stage manner, or directly predict interaction points in a one-stage procedure. However, the lack of spatial configurations or reasoning steps of two- or one- stage methods respectively limits their performance in such complex scenes.
To avoid this ambiguity, we propose a novel actor-centric framework. 
The main ideas are that when inferring interactions: 
%%%%%%%%%%%%%%%%%%%%%
%In the decision letter, the editor suggested to optimize this long sentence for easier understanding:
%%%%%%%%%%%%%%%%%%%%%
% 1) non-local features guided by binary masks generated from each detected human, \textit{i.e.}, actor, without the cropping procedure would be obtained, and then 2) every single actor is able to retrieve all objects of the entire image on the basis of these non-local features, and then evaluates the interactiveness independently.
1) the non-local features of the entire image guided by actor position are obtained to model the relationship between the actor and context, and then 2) we use an object branch to generate pixel-wise interaction area prediction, where the interaction area denotes the object central area. Moreover, we also use an actor branch to get interaction prediction of the actor and propose a novel composition strategy based on center-point indexing to generate the final HOI prediction.  
Thanks to the usage of the non-local features and the partly-coupled property of the human-objects composition strategy, our proposed framework can detect HOI more accurately especially for complex images.
Extensive experimental results show that
our method achieves the state-of-the-art on the challenging V-COCO and HICO-DET benchmarks and is more robust especially in multiple persons and/or objects scenes.
\end{abstract}


\end{frontmatter}

\section{Introduction}

Human-Object Interaction (HOI) detection \cite{chen2014predicting,  yao2010modeling, gao2020interactgan}, aiming to detect human, object and corresponding interaction between them 
from a given image, is a meaningful task serving as the fundamental step for many computer vision applications such as robotics \cite{aksoy2011learning,worgotter2013simple,argall2009survey} and activity analysis \cite{yang2013detection}. It is still a challenging task due to the inevitable ambiguities that come from the various visual relationships between multiple persons and objects occurring in the scene.

\begin{figure}[!htb]
	\centering
	\includegraphics[width=\linewidth]{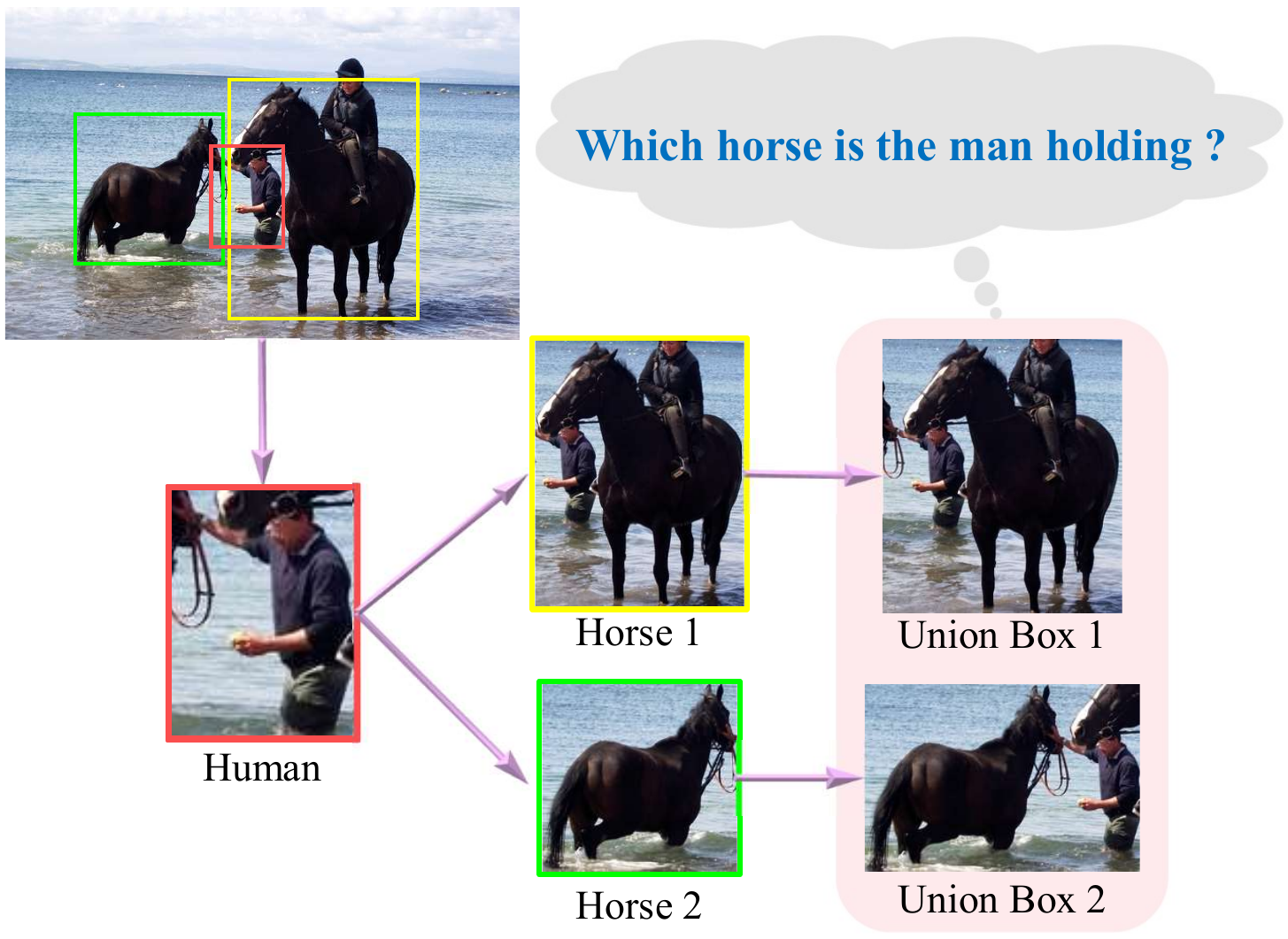}
	\caption{How to differentiate which horse is interacting with the person in the \textcolor{red}{red} box? The horse 1~(in the \textcolor{yellow}{yellow} box) is very close to the human, resulting in ambiguities between two interaction pairs with only cropped local features \cite{gkioxari2018detecting, hou2020visual}. This motivates us to consider multiple objects in the image simultaneously while predicting one person's interactions to avoid this ambiguity.}

	\label{fig:first}
\end{figure}

Given an image, most existing methods \cite{gkioxari2018detecting, chao2018learning,hou2020visual, xu2019learning, liu2020consnet} detect humans and objects first and, for each human-object pair, recognize the probable interactions with the local features cropped according to their bounding boxes. 
However, though such a pipeline works well for simple scenes when only appears few people, it may fail on complex scenes with ambiguities caused by multiple persons and objects, as shown in Fig.\ref{fig:first}. 
To tackle this issue, a few approaches \cite{gao2018ican, ulutan2020vsgnet, liao2020ppdm} try to utilize context information to complement local appearance to obtain a better detection result. Some of them \cite{gao2018ican,ulutan2020vsgnet,wang2019deep} seek to explore contextual representation as supportive information for the cropped local features, and improve detection results stably. However, they usually design an additional branch to extract contextual information by Global Average Pooling (GAP), which would cause an information loss, especially spatial configurations due to the pooling operation of all pixels. Thus such utilization of contextual information is limited. Other crop-free approaches \cite{liao2020ppdm,wang2020learning} formulate HOI detection as an interaction point detection problem and use one-stage strategy to exploit contextual information directly. Although these methods are quite effective in many cases, their highly-coupled property and the lack of reasoning steps~\cite{zhong2021glance} make the model hard to work well on complex images.

To overcome these issues, we present a novel partly-coupled actor-centric HOI detection framework. On the one hand, we formulate HOI detection task as a pixel-wise prediction problem in a one-stage like procedure, which avoids the information loss by GAP operations.
On the other hand, non-local features guided by binary masks generated from each detected human without the cropping procedure would be obtained. On the basis of these non-local features, every single human is able to evaluate the existence of interaction for all objects in the image. 
More specifically, we present an actor-centric model. In this model, we propose RGBM Generator module, which generates binary mask~(M) according to human bounding box and concatenate it with the original image~(RGB). The generated RGBM data is used as input of feature extraction backbone. 
Then, we regress interacting confidence map explicitly with a square mask supervision which could be inferred by original annotations directly. 
Finally, we propose an efficient post-processing procedure to generate HOI predictions for this brand-new framework, in which a composition strategy based on center-point indexing is used.
Thanks to the usage of the non-local features and the partly-coupled property of the human-objects composition strategy, our proposed framework can detect HOI more accurately especially for complex images.

To summarize, our contributions are four-folded:

\begin{itemize}
	\item We propose a novel actor-centric HOI detection framework to explore the relationships between one human and multiple objects, addressing the ambiguity issue across multiple interactions in the image. 
	\item We formulate the HOI detection task as a pixel-wise classification problem in a one-stage-like procedure with a proposed Weighted Cross Entropy Loss (WCEL), which could reduce the overfit of the boundary hop.
	\item We present an RGBM Generator module to provide an actor-centric guidance of the network, and design an efficient composition strategy to obtain the final scores by a scoremap indexing post-process.
	\item Extensive experimental results and ablation studies demonstrate the superiority of our approach in both quantitative and qualitative results in challenging HICO-DET and V-COCO benchmarks, especially for complex images.
\end{itemize}
\section{RELATED WORK}

Existing HOI detection methods can be roughly categorized into crop-based and crop-free fashions~\cite{zhong2021glance,zhang2021mining,ASNET,li2021improving}.

Crop-based HOI detection methods~\cite{gkioxari2018detecting, chao2018learning,hou2020visual, xu2019learning, liu2020consnet,gao2018ican} utilize Faster RCNN or Mask RCNN to generate human/object bounding box with the corresponding confidence in the first stage. And in the second stage, the interaction between human and object is detected by parsing the cropped features in the union box. Chao~\textit{et al.}~\cite{chao2018learning} propose a multi-stream framework including human-stream, object-stream and 
pairwise-stream to obtain verb score, and then the verb confidence is obtained by adding three scores from all streams. Gao~\textit{et al.}~\cite{gao2018ican,wang2019deep} enhance the feature in human/object stream through attention mechanism and obtain a significant improvement thanks to aggregating the global information.

However, their performance is limited because the spatial and contextual information are often lost during the cropping procedure, which might cause ambiguous situations, as shown in Fig.~\ref{fig:first}. 
In contrast, our method recognizes interaction in a crop-free manner. We leverage a global context extraction network instead of cropped feature fusion network to make sure every pixel is able to obtain global information. With the supervision of our proposed loss, each pixel is guided to output interaction score of the object/human it belongs to.

Recently, crop-free HOI detection methods\cite{wang2020learning, liao2020ppdm} have attracted increasing attention because previous works struggle with the problem of efficiency. Most crop-free methods extract global-aware features for better performance and take HOI as a key-point detection task. IP-Net~\cite{wang2020learning} detects the interactions between human-object pairs through Hourglass-104~\cite{newell2016stacked, law2018cornernet} and then associates human and object
proposal generated by FPN~\cite{lin2017feature} to obtain final HOI predictions. PPDM~\cite{liao2020ppdm} defines the localization for interaction. Human and object points are the center of
the detection boxes, and the interaction point is the midpoint of the human and object points. It utilizes a matching strategy in the post-processing, and makes the pipeline in one-stage.

Although the efficiency of one-stage methods is usually relatively satisfactory, the performance suffered since there might exist conflicts of interaction key-points when multiple persons or objects are shown. To avoid such conflicts, we innovatively constrain the network to predict the interaction at the central area of human/object.

\begin{figure*}
	\centering
	\includegraphics[width=\textwidth]{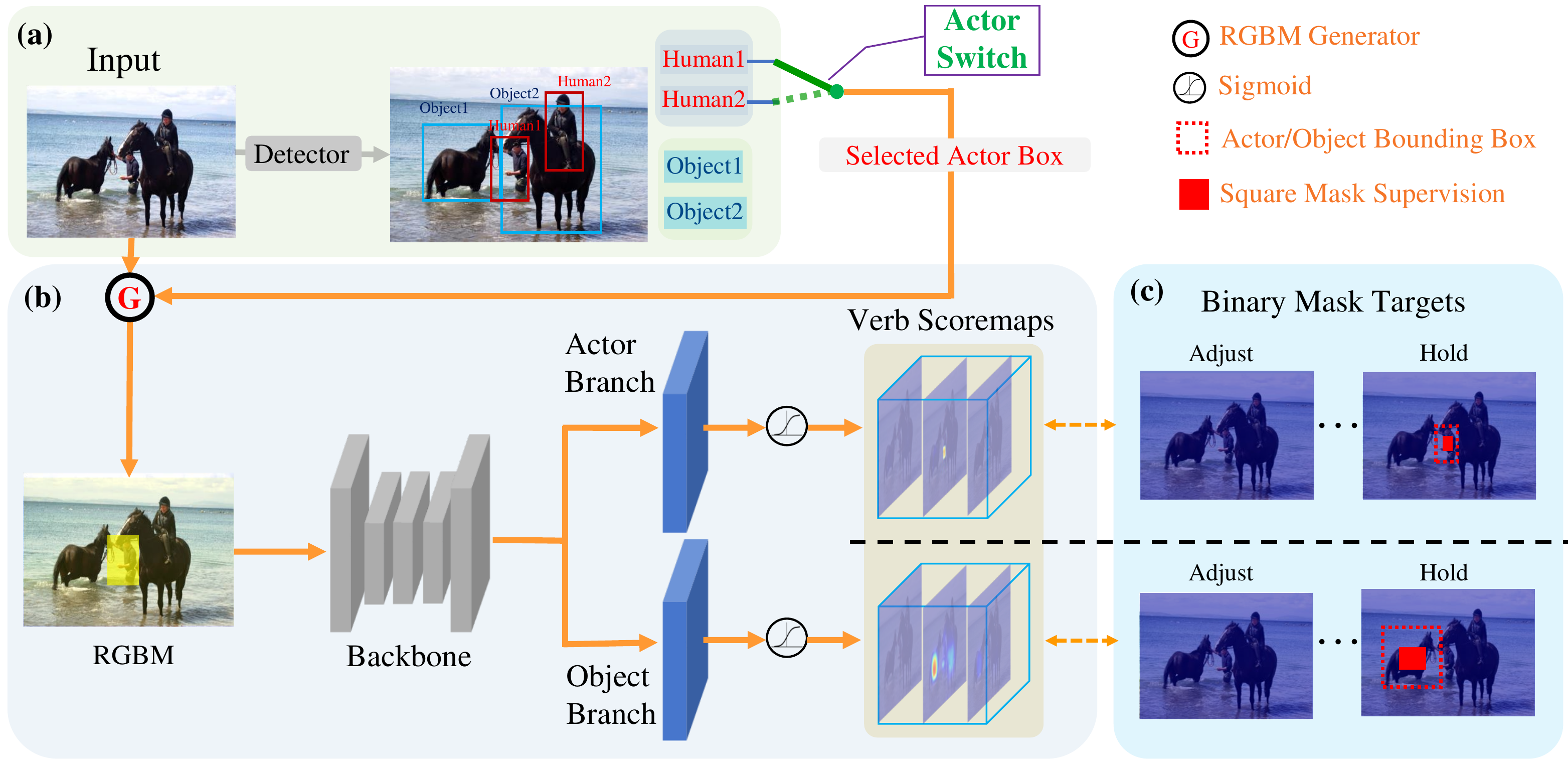}
	\caption{
		An overview of our framework for Human-Object Interaction Detection. We formulate the HOI detection task as a pixel-wise verb score prediction problem in two branches, \textit{i.e.}, Actor Branch and Object Branch. The input of the network is generated by an RGBM Generator based on the Actor Switch that selects each human in the image sequentially. 
		In order to force the network to predict interaction location and scores simultaneously, we design a strategy that generates square masks according to original annotations and adopt the masks as supervision.
		Guided by actor mask in the RGBM data, our framework could exploit the relationship between the actor and multiple objects in the image to generate accurate predictions. 
	}
	\label{fig:framework}
\end{figure*}
\section{METHODOLOGY}

Given an input image $I \in \mathbb{R}^{W \times H}$, the goal of HOI detection is detecting the $ \langle human, verb, object \rangle$ triplets, such as $ \langle human, hold, cup \rangle$, $ \langle human, ride, horse \rangle$. In this work, we denote the verb set as $\Psi=\{\psi_{i} \}_{i=1}^{K}$, where $K$ is the total number of verb categories.

\subsection{Overview}

Our proposed framework is built on a pre-trained object detector and focuses on inferring the triplets categories for each human in the image. As illustrated in Fig. \ref{fig:framework}, we firstly use an object detector to locate humans and objects, then traverse all detected humans by an Actor Switch module sequentially. The selected human is defined as actor and utilized to generate an RGBM four-channel data (described in the following chapter).
Then, a backbone \cite{yu2018deep, sun2019deep} is used to extract features, behind which the framework separates into two branches, namely Actor Branch and Object Branch. Actor Branch aims to predict pixel-wise verb scores of the current actor. 
When the actor is performing one interaction, the corresponding verb scores in the actor's central area is expected to be high. 
Similarly, for Object Branch, we also obtain per-pixel object verb predictions. For the expectation of the network to focus on the center location of these agents, we narrow the original actor/object box into a smaller binary square mask, which has been shown by the red area in Fig. \ref{fig:framework}. These red areas are generated as the targets of our network, thus we can cooperate the current actor verb score and the corresponding multiple objects verb scores from the two branches to predict the interaction triplets.

\begin{figure}[!htb]
	\centering
	\includegraphics[width=\linewidth]{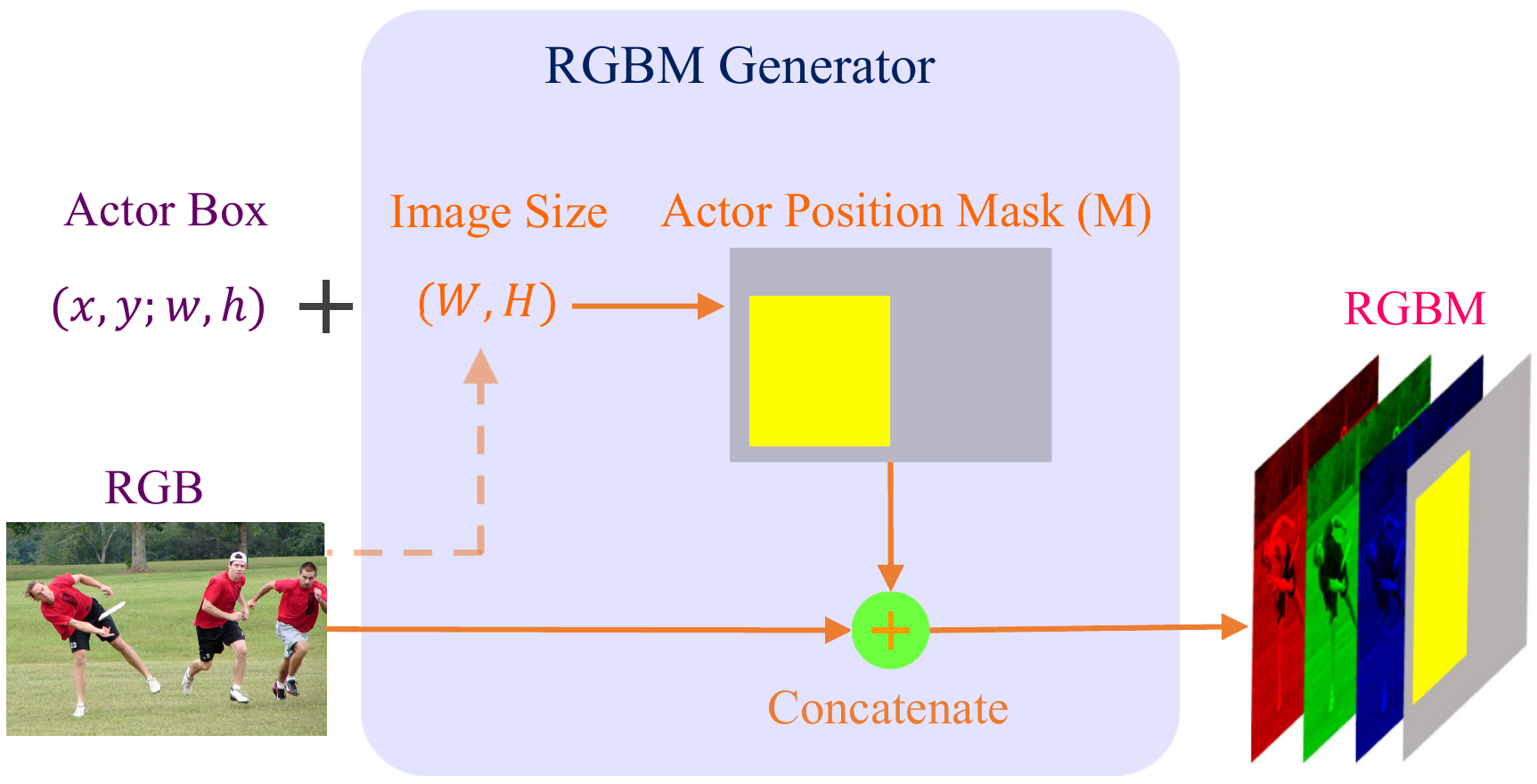}
	\caption{Illustration of RGBM data generation. We first generate a Actor Position Mask, \textit{i.e.}, $M \in \mathbb{R}^{W \times H \times 1}$, based on the image size of the RGB image and coordinates of the Actor box. Then we concatenate the mask with the original RGB image to obtain four-channel RGBM data as the inputs of our framework.}
	\label{fig:rgbm}
\end{figure}
\subsection{RGBM Generation and Feature Extraction}
\label{chap:RGBM}
We denote detected human set as $H=\{h_{i}\}_{i=1}^{M}$, where $h_{i}$ is a detected human and $M$ is the number of detected humans. Similarly, when $N$ objects is detected, we denote detected object set as $O=\{o_{i}\}_{i=1}^{N}$, where $o_{i}$ is a detected object and $N$ is the number of detected objects.

There are many ways to indicate human position. We choose to use an additional $W\times H$ mask map to avoid damaging the original image structure of the RGB channel. 

As is shown in Fig. \ref{fig:rgbm}, given a human box and an input image $I$, we generate a human position mask map with the same height and width with $I$. The area within the human box is set as 1 and other positions are set as 0. Then we concatenate the human position mask and origin image. We denote the generated 4-channel data as RGBM image.

In order to ensure every position on the image could utilize global context and local appearance to generate verb prediction, we adopt DLA \cite{yu2018deep} or HRNet \cite{sun2019deep} which are proposed to effectively aggregate global and local information, as our backbone. After feature extraction, the feature resolution is down from $W\times H$ to $W/d \times H/d$, where d is the down-sample stride. 

\subsection{Actor Branch and Object Branch}
The Actor Branch and Object Branch are two detection heads to generate pixel-wise verb prediction for actor and all objects separately. In Fig. \ref{fig:framework}, output size of each branch is  $W/d \times H/d \times (K+1)$ where $K$ is the number of annotated verb categories. We use the first $K$ channels to generate predictions for all verb categories. The last channel is used to generate prediction for an additional category which is named w/o-interaction. 

For Actor Branch, we request the network to generate a high score in the area around actor box center when an interaction happens. In practice, we use a box smaller than actor box as the central area which is denoted as $Z$. Furthermore, we denote the interaction set the actor participates in as $\Psi_{a}$. Note that if the actor is not interacting with any object, we set $\Psi_{a}=\{$w/o-interaction$\}$.
And we generate target illustrated in Fig. \ref{fig:framework} by
\begin{equation}
	f_{a}(x,y,c)=\left\{    
	\begin{array}{cc}
		1& (x,y) \in Z,  \Psi^{*}[c] \in \Psi_{a}\\ 
		0& otherwise 
	\end{array}
	\right. ,
	\label{equ:target}
\end{equation} 
where $(x, y)$ denotes a spatial position and $c$ denotes channel. $\Psi^{*}$ is expanded from origin verb set $\Psi$ by adding the additional category w/o-interaction. $\Psi^{*} [c] \in \Psi^{*}$ is the verb category channel $c$ represents.

For Object Branch, given $o_{i} \in O$, we denote the interaction set $o_{i}$ participates in with the actor as $\Psi_{o_{i}}$ and we set $\Psi_{o_{i}}=\{$w/o-interaction$\}$ if $o_{i}$ is not interacting with the actor . We also generate target for each object by Equ. \ref{equ:target} and denote the generated target is $f_{o_{i}}$. Note that the w/o-interaction category here means $o_i$ is not interacting with the actor. The target for Object Branch denoted as $f_o$ is calculated by 
\begin{equation}
	f_{o}=f_{o_1}\oplus f_{o_2} \cdots \oplus f_{o_N}, 
\end{equation} 

where $\oplus$ denotes an addition rule that retains the maximum for elements in the same position.
\subsection{Loss}
Instead of simply using Cross Entropy Loss, we propose a Weighted Cross Entropy Loss (WCEL) based on following design:

\noindent
\textbf{Hanning Weight:} 
Because there is no transition between box central area and other area, the net work may pay too much attention on learning a sharp hop at the boundary of box central area and this could cause overfit. To settle this, we want the position near the center point has a higher weight and position near the boundary has a lower weight. In this work, we use Hanning window to generate weight for the positive positions and negative positions individually. 
The two-dimensional Hanning window is defined as:
\begin{small}
	\begin{equation}
		\mathcal{H}(x,y,w,h)=\frac{1}{4}[1+cos(2\pi \frac{x}{w-1}][1+cos(2\pi \frac{y}{h-1}] ,
		\label{equ:hanning}
	\end{equation}
\end{small}
where $w$ and $h$ denote width and height of Hanning window and $(x,y)$ is a point within Hanning window.

Given a box $B$ whose width, height and center point are $w$, $h$ and $(x_{0},y_{0})$ separately, the proposed Hanning Weight $w_{han}$ for positions within $B$ is calculated by 
\begin{small}
	\begin{equation}
		w_{han}(x,y,c)\!=\!\left\{\!\! 
		\begin{array}{c}
			\mathcal{H}(x\!-\!x_{0},y\!-\!y_{0},w,h) \quad \quad \quad \!\!\! f(x,y,c)\!=\!1\\ 
			1\!-\!\mathcal{H}(x\!-\!x_{0},y\!-\!y_{0},w,h) \quad f(x,y,c)\!=\!0 
		\end{array}
		\right..
		\label{equ:hanning_weight}
	\end{equation}
\end{small}
And we set $w_{han}(x,y,c)=1$ if $(x,y)$ is not within $B$. 

\noindent
\textbf{Scale Weight:}
Because the size of the central area is relevant to box size, if we simply use classification loss for every spatial position, bigger box will have bigger area to generate loss and smaller ones has smaller area to generate loss. Such imbalance of supervision area will cause smaller object/actor to become hard to optimize. To solve this problem, we give different area corresponding wights according to box position and size. We define this kind of weight as Scale Weight and express it as $w_{scale}$. Take the above box $B$ for example, the Scale Weight of position within $B$ is calculated by:

\begin{equation}
	w_{scale}(x,y,c)= min(10, \lambda_{s} max(W,H)/max(w,h)),
	\label{equ:scale_weight}
\end{equation}

where $\lambda_{s}$ is a super parameter and we set it as $0.5$ in our work. Besides, we set the upper limit of $w_{scale}$ as $10$ to prevent the weight of small box greater than that of background too much. Similar to Hanning Weight, we set $w_{scale}(x,y,c)=1$ if $(x,y)$ is not within $B$.

\noindent
\textbf{Weighted Cross Entropy Loss:}
Assume the output of network is $\hat{f} \in \mathbb{R}^{W \times H \times (K+1)}$ and corresponding target, Hanning Weight and Scaled Weight is $y$, $w_{han}$, $w_{scale}$ separately. The loss for $\hat{f}$, denoted as $\mathcal{L}$, is calculated by
\begin{equation}
	\begin{aligned}
		\mathcal{L}\!=\!\sum_{x,y,c}w_{han}&(x,y,c)w_{scale}(x,y,c)[f(x,y,c)log(\hat{f}(x,y,c))\\ &+(1-f(x,y,c))log(1-\hat{f}(x,y,c))].
	\end{aligned}
	\label{equ:loss}
\end{equation} 

Assume $\hat{f}_{a}$ and $\hat{f}_{o}$ are outputs of Actor Branch and Object Branch respectively. We denote the loss of Actor Branch and Object Branch as $\mathcal{L}_{a}$ and $\mathcal{L}_{o}$, and they can be calculated by Equ. \ref{equ:loss}. 

Total loss of our framework is denoted as 

\begin{equation}
	\textbf{L}= \lambda_{a}\mathcal{L}_{a}+ \lambda_{o}\mathcal{L}_{o},
	\label{equ:loss_branch}
\end{equation}
where $\lambda_{a}$ and $\lambda_{o}$ are two super parameters to balance the losses of Object Branch and Human Branch. In our work, we set $\lambda_{a}=\lambda_{o}=1$.

\begin{figure}
	\centering
	\includegraphics[width=\linewidth]{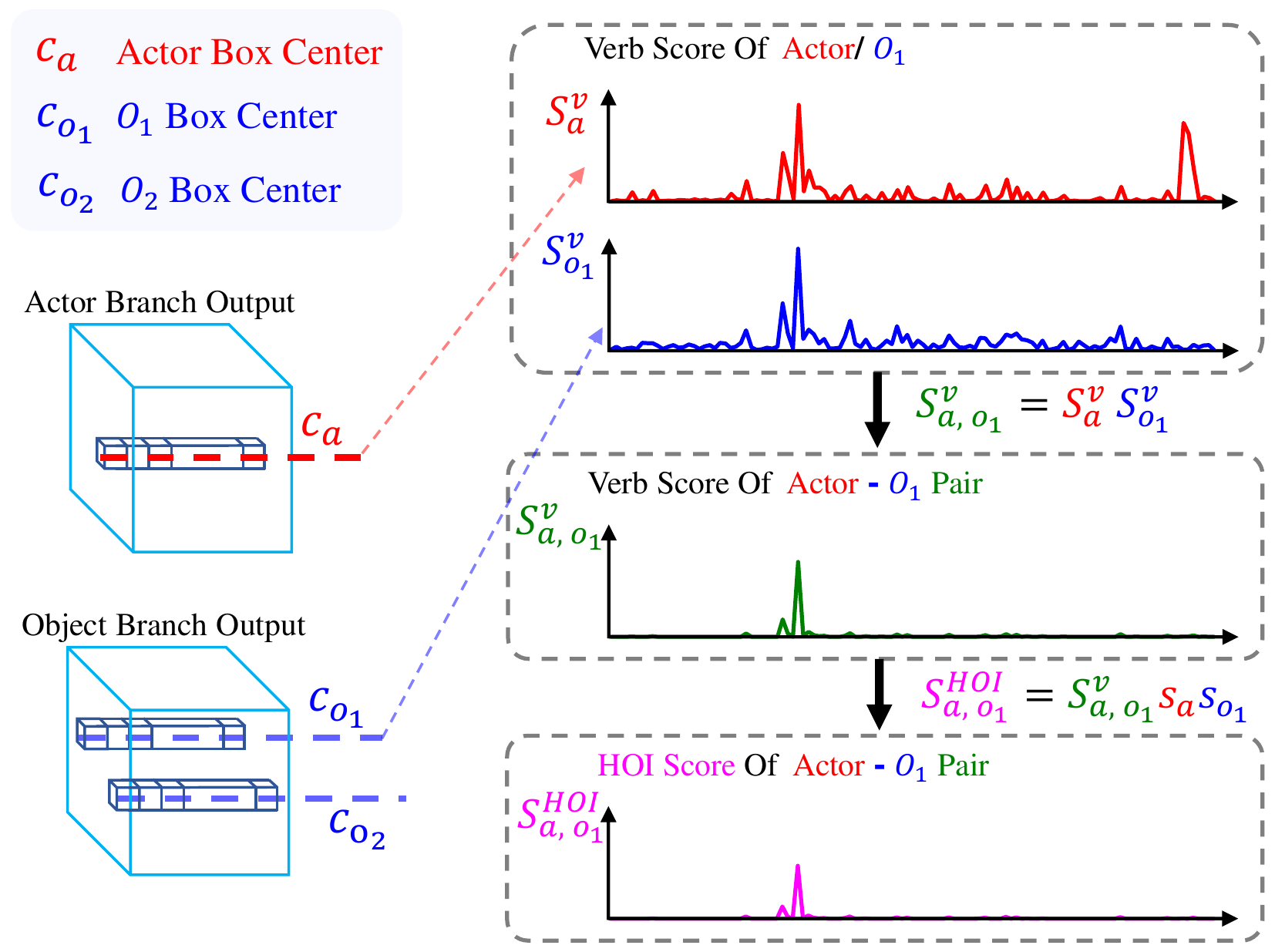}
	\caption{Illustration of final score predictions. We first calculate verb scores of the actor and objects by indexing on two branches output based on the center point coordinates of their bounding boxes. Then the final HOI prediction can be obtained by multiplying the verb scores and the detection scores.}
	
	\label{fig:inference}
\end{figure}

\subsection{Inference}
During inference, we use the pipeline in Fig. \ref{fig:framework} to generate pixel-wise verb predictions first. 
Then, we use a simple post-process strategy  illustrated in Fig. \ref{fig:inference} to generate final HOI scores. Given the selected actor and object set $O=\left\{o_i\right\}_{i=1}^{N}$ provided by detector, we obtain object center point set $\mathcal{C}_{o}=\{c_{o_{i}}\}_{i=1}^{N}$ and  actor center point $c_{a}$. Then, we use $c_{a}$ to index actor verb score $S_{a}^{v}$. Simultaneously, we use $c_{o_{i}}$ to index object verb scores $S_{o_{i}}^{v}$ for each $o_{i}\in O$. Next we calculate verb score $S_{a,o_{i}}^{v}$ for an actor-$o_i$ pair by  

\begin{equation}
	S_{a,o}^{v} = S_{a}^{v} \cdot S_{o}^{v}.
\end{equation} 
And the HOI score $S_{a,o}^{hoi}$ of actor-$o_i$ pair is denoted as
\begin{equation}
	S_{a,o_{i}}^{hoi} = S_{a,o_{i}}^{v} \cdot s_{a} \cdot s_{o_{i}},
\end{equation} 
where $s_{o_{i}}$ is the object confidence score provided by object detector. $s_{a}$ is the confidence score of the actor (\textit{i.e.}, selected human) and is also provided by object detector.

\section{Experiments}
\subsection{Datasets and Metrics}
\noindent
\textbf{Datasets} We verify the performance of our method on HICO-DET~\cite{chao2018learning}, V-COCO~\cite{gupta2015visual} and HOI-A~\cite{liao2020ppdm}.  
HICO-DET contains 38118 images for training and 9658 images for testing. 
There are 80 object categories as same as MS-COCO~\cite{lin2014microsoft} and 117 verb categories. The object and verb categories composite 600 HOI categories for HOI detection. 
V-COCO is a smaller dataset which is derived from MS-COCO dataset. It contains 5400 images in the trainval dataset and 4946 images in the test dataset. Each person has annotations for 29 action categories which contains 25 HOI categories and 4 body motions.
HOI-A is a recently proposed dataset containing 11 kinds of objects and 10 action categories in general scenes.

\noindent
\textbf{Metrics} Following previous works, we adopt the mean average precision~(mAP) to evaluate our method. For HICO-DET, AP is computed on per HOI class and two setting are adopted, \textit{i.e.}, ``Known Object'' and ``Default''. In ``Known Object'' setting, we evaluate only on the images containing the annotated object category for each HOI category, and in ``Default'' setting, we evaluate on full test images. In each setting, we report the mAP over three types, \textit{i.e.}, Rare, Non-Rare and Full, which are defined according to number of training instances. 
For V-COCO, three settings are adopted. In $AP_{agent}$, the true positive only focus on the pair $\left\langle subject, verb\right\rangle$. 
While $AP_{role}^{S1}$ and $AP_{role}^{S2}$ requires object should be also correctly located.
For the cases when there is no object, in $AP_{role}^{S1}$ a
prediction is correct if the corresponding bounding box for
the object is empty and in $AP_{role}^{S2}$ the bounding box of
the object is not considered. 
For HOI-A, AP is computed on per verb class.

\begin{table}
	\centering
	\setlength{\tabcolsep}{10pt}
	% \label{tab:vcoco}
	\begin{tabular}{llccc}
		\hline \hline
		Methods  &  $AP_{agent}$  & $AP_{role}^{S1}$ & $AP_{role}^{S2}$ \\ \hline
		InteractNet~\cite{gkioxari2018detecting}  & 69.2 &40.0  &-  \\
		GPNN~\cite{qi2018learning}  & - &44.5 & 42.8 \\ 
		iCAN~\cite{gao2018ican} &   - &45.3  & 52.4 \\ 
		RPNN~\cite{zhou2019relation} &  - &47.5 & - \\
		VCL~\cite{hou2020visual} &  - & 48.3&  -\\
		
		TIN${}^*$~\cite{li2019tin} &   -& 48.7   &54.2  \\
		
		Zhou et al.~\cite{zhou2020cascaded} & -  &  48.9& \\
		TIK~\cite{li2019transferable} &-&48.7&-\\
		
		PastaNet${}^*$~\cite{li2020pastanet} &  - & 51.0 & 57.5 \\
		
		DRG${}^*$~\cite{gao2020drg} & -  &  51.0 & \\
		
		Wei et al.~\cite{feng2019turbo} & 70.3 &  42.0 &  -\\
		ACP${}^*$~\cite{eccv2020actioncoprior} & -  &  53.2 & \\
		IDN~\cite{li2020hoi} & -  &\textbf{53.3}   & 60.3 \\
		% \textbf{Ours} & ResNet-50  &  &  & 52.8 \\ 
		\hline 
		
		Ours  & \textbf{73.47} & 51.67 & \textbf{61.75} \\ 
		
		\hline \hline
	\end{tabular}
	\caption{Result on V-COCO. Character ${}^*$ indicates that external knowledge is used. 
	}
	\vspace{-0.2cm}
	\label{tab:vcoco}
\end{table}

\subsection{Implementation Details}

\noindent
\textbf{Object Detector.} 

We use Faster R-CNN~\cite{ren2016faster} as the object detector. In experiments, we adopt the pre-trained model released by MMDetection~\cite{chen2019mmdetection} to localize persons and objects. 
The NMS threshold is set as $0.05$ and top-100 predictions are used for later processing.
We also test our method on HICO-DET by utilizing the detection results released by DRG\cite{gao2020drg} with the model that finetuned on HICO-DET. When using the detection results of DRG, we set the threshold of both human and object score to 0.2.

\begin{table*}[!htbp]
	\centering
	\setlength{\tabcolsep}{4pt}
	\begin{tabular}{lc|ccc|ccc}
		\hline \hline
		& \multicolumn{1}{l}{}&  \multicolumn{3}{|c}{Default}                                                                         & \multicolumn{3}{|c}{Known Object}                     \\
		Methods                                  & External Knowledge     & Full                  & Rare       & Non-Rare         & Full         & Rare          & Non-Rare \\ \hline
		iCAN~\cite{gao2018ican}                  &                       & 14.84                  & 10.45               & 16.15                &     16.26      &     11.33           &    17.73     \\
		Wang~\textit{et al.}~\cite{wang2019deep}      &                       &16.24                         &11.16              & 17.75                 &17.73          &12.78              & 19.21     \\
		PMFNet~\cite{wan2019pose}                    & P                     & 17.46                  & 15.65                 & 18.00               &      20.34     &     17.47           &     21.20       \\
		No-Frills~\cite{gupta2019no}                    & L                     & 17.18                  & 12.17                & 18.68               &    -            &         -       &    -               \\
		TIN~\cite{li2019tin}                          & P                     & 17.22                & 13.51                  & 18.32              &   19.38         &      15.38          &    20.57      \\
		CHG~\cite{eccv2020hetegraph}                    &                       & 17.57                 & 16.85                  & 17.78              &  21.00          &    20.74            &    21.08       \\
		UnionDet~\cite{kim2020uniondet}            &                       & 17.58                  & 11.52                 & 19.33              &  19.76           &  14.68              &     21.27        \\ 
		Peyre~\textit{et al.}~\cite{peyre2019detecting}           &         L             & 19.40                 & 14.63                 & 20.87               &     -           &      -          &     -              \\
		VSGNet~\cite{ulutan2020vsgnet}                    &                      & 19.80                   & 16.05                & 20.91               &     -           &     -           &      -             \\
		FCMNet~\cite{eccv2020keycues}                    & P+L                   & 20.41                 & 17.34                  & 21.56              &   22.04         &       18.97         &    23.12       \\
		ACP~\cite{eccv2020actioncoprior}                 & P+L                   & 20.59                 & 15.92                 & 21.98               &        -        &    -            &    -               \\

		Bansal~\textit{et al.}~\cite{bansal2020detecting}      &    L                  & 21.96                 & 16.43                  & 23.62             &      -          &           -     &     -              \\

		DRG~\cite{gao2020drg}                &     L                        & 24.53                 & 19.47                  & 26.04              &     27.98       &   23.11            &   29.43         \\

		IDN~\cite{li2020hoi}               &                       & 26.29                 &\textbf{22.61}           & 27.39              & 28.24            &\textbf{24.47}               & 29.37       \\ 
		ConsNet-F~\cite{liu2020consnet}               &  L                   & 24.39                 & 17.10                  & 26.56               &  -          &   -             &           - \\ 
		IPNet~\cite{wang2020learning}                      &                    & 19.56                 & 12.79                  & 21.58              &    22.05         &      15.77          &     23.92        \\
		PPDM~\cite{liao2020ppdm}                     &                       & 21.73                 & 13.78                 & 24.10               &  24.58          &   16.65             &           26.84 \\ 
		HOTR~\cite{kim2021hotr}         &                                       & 25.10         &17.34                  &27.42               &  -          &   -             &         - \\ 
		Zou~\textit{et al.}~\cite{zou2021end}         &                       & 26.61                 & 19.15                 & 28.84               &  29.13          &   20.98             &          31.57 \\ 
		
		\hline 
		\textbf{Ours}                                   &                      & \textbf{27.39}         & 21.34                   &\textbf{29.20}       & \textbf{30.87}  &24.20             &  \textbf{32.87}     \\ 
		
		\hline \hline
	\end{tabular}
	\caption{Results on HICO-DET. 
		For External Knowledge, ``P'' indicates human pose and ``L'' indicates linguistic knowledge.	}
	\vspace{-0.2cm}
	\label{tab:hico}
\end{table*}

\noindent
\textbf{Annotation Arrangement.} For the convenience of training interaction classification network, we rewrite the training annotation: for each annotated person $p$, we record box of $p$, all objects interacting with $p$ and corresponding verbs. 
For a detected person $p_{0}$, 
if there exists an annotated person that has IoU over 0.5 with $p$, we assign object and verb annotation of the person to $p_{0}$,  otherwise we define $p_{0}$ as a negative example. Usually, the negative examples are several times of the positive ones. To tackle the imbalance of examples, we set the ratio between positives and negatives as $1:1$  by randomly dropping some negative examples.

\noindent
\textbf{Feature Extractor.} 
We use HRNet-W32 \cite{wang2019deep} and DLA-34\cite{yu2018deep} as our backbones for interaction classification and evaluating the effectiveness of our proposed idea. 
  HRNet was initially proposed for human pose estimation which requires the network to extract local and global features for effective keypoint detection. The HRNet-W32 used in this work is a relatively lightweight edition compared to HRNet-W48 and HRNet-W64. DLA is a more lightweight network, which is proposed to fuse information across layers of network, and it can also make sure every position in the output layer obtain information across scales. In our experiments, we initialize HRNet-W32 and DLA-34 with weights pre-trained on ImageNet~\cite{russakovsky2015imagenet}.
  
\noindent
\textbf{Other Parameters.} 
The bounding box of the supervised area illustrated in Fig.~\ref{fig:framework} is scaled from human/object box with a ratio of 0.3.
During training, we use ADAM~\cite{kingma2014adam} as the optimizer and train the model for 12 epochs with learning rate $1.5\times 10^{-5}$. Our experiments are all conducted on two Nvidia GeForce RTX 2080Ti GPUs.

\begin{table}[!htb]
	\centering
	\setlength{\tabcolsep}{15pt}
	% \label{tab:vcoco}
	\begin{tabular}{ll}
		\hline \hline
		Methods& mAP($\%$)   \\ \hline
		
		C-HOI~\cite{zhou2020cascaded} & 66.04  \\
		iCAN~\cite{gao2018ican} &44.23\\ 
		TIN~\cite{li2019tin} &48.64\\
		PPDM~\cite{liao2020ppdm}  &71.23\\
		Ours  &\textbf{74.56} \\ 
		
		\hline \hline
	\end{tabular}
	\caption{Result on HOI-A test set. 
	}
	\vspace{-0.5cm}
	\label{tab:hoia}
\end{table}

\subsection{Evaluation}
We conduct the quantitative experiments on V-COCO, HICO-DET and HOI-A to demonstrate the effectiveness of our method. The results are shown in Tab.~\ref{tab:vcoco}, Tab.~\ref{tab:hico} and Tab.~\ref{tab:hoia} in comparison with other classical methods. 
% Our results are obtained without using any external knowledge.

For V-COCO, we report our result in terms of $AP_{agent}$, $AP_{role}^{S_{1}}$ and $AP_{role}^{S_{2}}$. Our method outperforms state-of-the-art method by $3.04\%$ and $1.45\%$ on $AP_{agent}$ and $AP_{role}^{S_{2}}$ respectively. This shows that our method could not only achieve excellent HOI detection performance, but also promote human action analysis.

For HICO-DET, we report result on Default and Known Object settings. And for each setting, the results on Full, Rare and Non-Rare subset are shown in Tab.~\ref{tab:hico}. Although some methods~\cite{wan2019pose,gupta2019no,li2019tin,peyre2019detecting,eccv2020keycues,eccv2020actioncoprior,gao2020drg,liu2020consnet} adopt external human pose or linguistic knowledge or both to help HOI detection, our method outperforms them by a large margin especially on Full and Non-Rare set. In addition, compared to recently proposed global-context-aware methods~\cite{gao2018ican, wang2019deep,wang2020learning,liao2020ppdm,zou2021end}, our method outperforms them on all subsets. 

However, we only achieve competitive performance on $AP_{role}^{S_{1}}$ of V-COCO and Rare set of HICO-DET compaerd to IDN. Considering V-COCO is a smaller dataset than HICO-DET and Rare set of HICO-DET only has a few samples for each category, it means that our method needs relatively more samples to train a powerful model.

To show the adaptability of different datasets, we also test our method on HOI-A dataset. As is shown in Tab.~\ref{tab:hoia}, we achieve 3.33 mAP improvement compared to state-of-the-art PPDM. The result shows that our method can be applied to a variety of scenarios with Human-Object Interactions.

Following InteractNet~\cite{gkioxari2018detecting}, we also report Role AP for each interaction category and Average Role AP of all categories on V-COCO~\cite{gupta2015visual}. The result is shown in Tab.~\ref{tab:v-coco}. Our result outperforms result reported in~\cite{gkioxari2018detecting} by a large margin. Besides, through column 2-3 and column 4-5 we find that our method get $8.14\%$ and $7.09\%$ Average Role AP improvement on scenario\_1 and scenario\_2 when using ground truth (GT) human/object boxes instead of human/object boxes detected by Faster R-CNN. Thus, HOI detection with our framework could be further promoted once a better detector is adopted.

\begin{table}[!htb]
	\centering
	\setlength{\tabcolsep}{1mm}
	\begin{tabular}{l|cc|cc}
		\hline
% 		\multirow{2}{*}{action}
		& \multicolumn{2}{c|}{scenario\_1} & \multicolumn{2}{c}{scenario\_2} \\
		\cline{2-5}
		 action& \multicolumn{1}{c}{Detector} & \multicolumn{1}{c|}{GT} & \multicolumn{1}{c}{Detector} & \multicolumn{1}{c}{GT} \\
		\hline
		hold-obj & 36.52 & 41.83 & 55.20  & 62.21 \\
		
		sit-instr & 29.25 & 40.25 & 55.58 & 63.93 \\
		
		ride-instr & 71.86 & 79.41 & 77.68 & 83.42 \\
		
		look-obj & 38.82 & 40.23 & 51.59 & 53.64 \\
		
		hit-instr & 77.11 & 83.80  & 83.37 & 91.07 \\
		
		hit-obj & 49.24 & 55.27 & 56.31 & 60.49 \\
		
		eat-obj & 43.80  & 50.49 & 70.35 & 70.26 \\
		
		eat-instr & 6.68  & 13.22 & 29.31 & 32.50 \\
		
		jump-instr & 56.97 & 60.85 & 60.08 & 61.33 \\
		
		lay-instr & 32.44 & 50.59 & 40.07 & 55.64 \\
		
		talk\_on\_phone-instr & 58.77 & 78.20  & 66.05 & 90.66 \\
		
		carry-obj & 39.24 & 41.72 & 43.58 & 45.75 \\
		
		throw-obj & 49.60  & 51.13 & 53.82 & 55.60 \\
		
		catch-obj & 46.74 & 46.22 & 57.19 & 57.21 \\
		
		cut-instr & 47.70  & 66.52 & 55.39 & 75.36 \\
		
		cut-obj & 39.96 & 43.83 & 54.21 & 59.43 \\
		
		work\_on\_computer-instr & 63.75 & 73.79 & 70.51 & 79.67 \\
		
		ski-instr & 52.76 & 68.56 & 65.57 & 80.27 \\
		
		surf-instr & 81.31 & 89.90  & 84.90  & 93.44 \\
		
		skateboard-instr & 89.27 & 93.65 & 92.89 & 96.71 \\
		
		drink-instr & 34.91 & 43.19 & 37.10  & 43.19 \\
		
		kick-obj & 75.98 & 76.14 & 85.39 & 85.30 \\
		
		read-obj & 37.64 & 57.31 & 49.69 & 62.10 \\
		
		snowboard-instr & 79.84 & 89.32 & 86.11 & 93.02 \\
		\hline
		Average Role AP & 51.67 & 59.81 & 61.75 & 68.84 \\
		\hline 
	\end{tabular}%
	\caption{Results of each verb category on V-COCO test set. We report the result on scenario\_1, and scenario\_2. ``Detector'' means using detected human/object boxes. GT means using ground truth human/object boxes.
	}
	\label{tab:v-coco}%
\end{table}%

\begin{table}[!htb]
\centering
	\subfloat[Effectiveness of RGBM Generator. We try different inputs of verb prediction network (Fig.~\ref{fig:framework}b). ``RGB'' means using the original image as input. ``RGB+255'' means add 255 to the intensity of pixels that lie inside actor bounding box.
	``RGBM'' means using proposed RGBM data as input.
	\label{tab:rgbm-generator}]{
		\setlength{\tabcolsep}{4.3mm}
		\begin{tabular}{c|ccc} 
			\hline
			Input             & Full  & Rare  & Non-Rare \\
			\hline % 中边框
			RGB               & 21.88 &16.84 &23.39 \\
			RGB+255         & 24.51 &18.77 &26.23    \\
		
			RGBM         &26.18&19.88&29.06  \\
			\hline
		\end{tabular}
	}

	\subfloat[Impact of Actor Branch. The Baseline only uses Object Branch for training and testing. Then we add Actor Branch during training procedure and further fuse prediction of Actor Branch and Object Branch when testing. 
	The results indicate that just learning with Actor Branch could promote the performance. Fusing prediction of both branches could bring further improvement. 
	%We use OOPO with Object Branch as base model(first row) and add Human Branch for Training and Testing.  
	\label{tab:actor-branch}]{
		\setlength{\tabcolsep}{2.2mm}
		\begin{tabular}{c|cc|ccc} % 创建一个5列的表格
			\hline
			&\multicolumn{2}{|c|}{Actor-Branch}&&&\\
% 			\cline{2-3}
			Baseline&Train          & Test      & Full  & Rare  & Non-Rare \\
			\hline % 中边框
		$\checkmark$&   &            & 25.60 &18.48  &27.73 \\
		$\checkmark$&  $\checkmark$  &  & 26.44 &18.97  & 28.68    \\
		$\checkmark$&$\checkmark$  &$\checkmark$&27.39 &21.34  &29.20  \\
			\hline
		\end{tabular}
	}
	\
	\subfloat[Effectiveness of Additional Category. Baseline uses the original verb categories as supervision for both branches. $A^{+}$ and $O^{+}$ means adding w/o-interaction category as supervision to Actor Branch and Object Branch respectively. We achieve best improvement when just adding w/o-interaction category to Object Branch.
	\label{tab:class}]{
		\setlength{\tabcolsep}{2.8mm}
		\begin{tabular}{ccc|ccc}
			
			\hline
			Baseline    & $A^{+}$         & $O^{+}$         & Full      & Rare      & Non-Rare\\
			\hline
			$\checkmark$ &              &               & 26.92     &21.18            &28.64   \\
			$\checkmark$& $\checkmark$  &               & 27.33     & 21.33     &29.13    \\
		    $\checkmark$&               & $\checkmark$  & 27.39      & 21.34            & 29.20  \\
			$\checkmark$& $\checkmark$  & $\checkmark$   & 27.07      &20.42  &29.07 \\
			
			\hline
			
	\end{tabular}}\vspace{1mm}
	\
	\subfloat[Effectiveness of Weighted Cross Entropy Loss. Standard binary cross entropy loss is used by default. 
	We show the results when using Hanning Weight and Scale Weight alone as well as using both together.
% It's obvious that both weights could bring significant improvement to Rare set.
	It's obvious that both loss weights could help to improve performance and the performance could be further improved when they are used together.
	\label{tab:loss}]{
		% 		\tablestyle{2pt}{1.05}
		\setlength{\tabcolsep}{1.1mm}
		\begin{tabular}{cc|ccc}
			\hline
			Hanning Weight&Scale Weight & Full & Rare & Non-Rare \\
			\hline
			&              &25.30  &18.41  &27.36\\
			$\checkmark$&               &25.61  &19.36 &27.53 \\
			&$\checkmark$   &25.45  &19.10 &27.41\\
			$\checkmark$&$\checkmark$   &25.83  &19.79 &27.60\\
			\hline 
	\end{tabular}}\vspace{1mm}

	\subfloat[Results on different backbones.
% 	Ablation studies on HICO-DET. 
	Our method outperforms existing methods when using the same backbone.  We achieves best results when using HRNet-W32. 
% 	which is a strong global information aggregation backbone.
	\label{tab:back}]{
		% 		\tablestyle{2pt}{1.05}
		\setlength{\tabcolsep}{1.4mm}
		\begin{tabular}{ll|ccc}
			\hline 
    		Methods&Backbone  &  Full  & Rare & Non-Rare \\ \hline
    		VCL~\cite{hou2020visual}  &Res-50&23.63& 17.21 &25.55\\
    		PastaNet~\cite{li2020pastanet} &Res-50& 22.65&\textbf{21.17}&23.09 \\
    		
    		Ours& Res-50  & \textbf{24.58}&17.71&\textbf{26.63} \\
    		\hline
    		PPDM~\cite{liao2020ppdm} &DLA34 & 20.29&13.06&22.45  \\
    		Ours &DLA34  &\textbf{26.18}&\textbf{19.88}&\textbf{29.06}\\ 
    		\hline
    		Ours  &HRNet-W32 &\textbf{27.39}& \textbf{21.34}& \textbf{29.20} \\ 
    		
    		\hline 
	\end{tabular}}\vspace{1mm}

	\caption{Ablation studies on HICO-DET. 
	}
	\label{tab:ablations}
	\vspace{1mm}
\end{table}

\noindent

\begin{figure*}
	\centering
	\includegraphics[width=\linewidth]{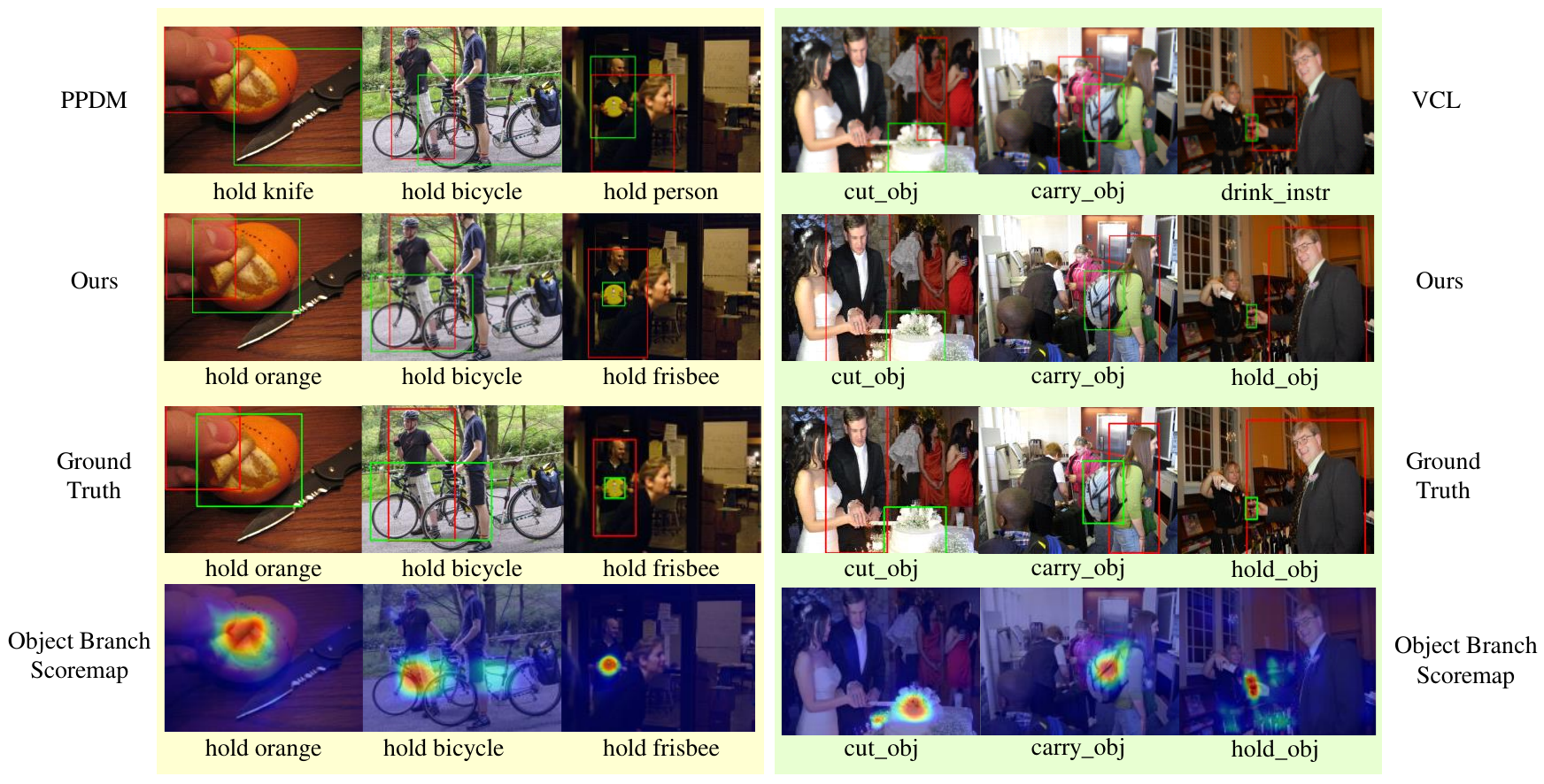}
	\caption{Visualization comparison results of our method and PPDM~\cite{liao2020ppdm} and VCL~\cite{hou2020visual} on some complex scenes from HICO-DET and V-COCO datasets. From top to bottom: The results of PPDM and VCL, our detection result, ground truth, and object scoremap predicted by our Object Branch. Clearly, in each specific HOI detection, high scores would be regressed for the human and its interacted objects. Thus, our approach obtains better performance on these complex cases, which validates the effectiveness of our framework to handle ambiguity issues.}
	\vspace{1.2mm}
	\label{fig:visual}
\end{figure*}

\subsection{Ablation Study}

\noindent
\textbf{RGBM Generator}
In Tab.~\ref{tab:rgbm-generator}, we conduct an ablation study to show the effectiveness of  RGBM generator. 
``RGB'' represents not indicating actor position at all. ``RGB +255'' represents indicating actor position by changing image structure of ``RGB'' channels. ``RGBM'' represents indicating actor position in the additional channel as proposed in the section of METHODOLOGY. We see a successive improvement from Line 1 to Line 3. This proves that ``RGBM'' is an effective way to guide interaction reasoning.

\noindent
\textbf{Actor Branch}
To validate the necessity of our Actor Branch, we conduct an ablation study shown in Tab.~\ref{tab:actor-branch}. We start from a baseline without actor branch in the training and testing process, and obtains 25.60 mAP in full subsets. While we add the actor branch on the training procedure and testing procedure, the performance obtains an improvement with 0.84 and 1.79 respectively. This proves that Actor Branch can guide the Object Branch to learn better. And directly fusing prediction of Actor Branch and Object Branch can further improve performance.

\noindent
\textbf{Additional Category}
We further evaluate the effectiveness of the additional category, \textit{i.e.}, w/o-interaction category% which has proposed in ``Actor Branch and Object Branch'' in Section 3.2
. This category is considered to add an explicit background category to split the interaction verb scores and non-interaction verb scores(though Non-interaction category is included in HICO-DET, it means that no interaction appears on the image). We add it to Actor Branch and Object Branch successively and obtains a slight improvement of 0.39 and 0.47. But when both branches are added simultaneously, the performance achieves a bit lower due to the difficult joint optimization of both branches.

\noindent
\textbf{Weighted Cross Entropy Loss}
The proposed weighted cross entropy loss consists of two types of weights, \textit{i.e.}, Hanning Weight, and Scale Weight. To validate their effectiveness, we conduct an ablation study as shown in Tab.~\ref{tab:loss}. We can see both loss weights could help to improve performance and the performance could be further improved when they are used together. 

\noindent
\textbf{Backbone}
To further verify the robustness of the proposed framework, we conduct experiments as shown in Tab.~\ref{tab:back} to compare with existing CNN-Based methods when using different backbones. We compare with VCL~\cite{hou2020visual}, PastaNet~\cite{li2020pastanet} and PPDM. VCL and PastaNet are both crop-based methods. Besides, PastaNet uses human pose and linguistic knowledge to guide interaction prediction. PPDM is one of the first methods that use non-crop features to predict interactions. 

Results in line 1-3 and line 4-6 show that our method brings significant improvements when using the same backbone~(no matter Res50 or DLA34), except for on Rare set compared to PastaNet~(which is with the help of the external knowledge for improving the performance on Rare set). We further achieve state-of-the-art results when using HRNet-W32 (line 5). The results show the adaptability of our framework to different backbones.
\begin{table*}[!htb]
\centering
	\subfloat[The results (mAP) on subsets of HICO-DET.
	(*\%$\uparrow$) denotes the increasing percentage we achieve compared to PPDM and VCL. 
	\label{tab:subset-table}]{
		\centering
	\setlength{\tabcolsep}{2.5mm}
	
	\begin{tabular}{l|cccc}
		\hline
		& SH-SO   & MH-SO    & SH-MO    & MH-MO \\
		\hline
	
		PPDM  & 29.03\textbf{(15\%$\uparrow$)} &17.17\textbf{(44\%$\uparrow$)}&  15.94\textbf{(55\%$\uparrow$)}&13.06\textbf{(41\%$\uparrow$)}  \\
		\hline
		VCL   & 24.78\textbf{(34\%$\uparrow$)} & 17.14\textbf{(45\%$\uparrow$)} & 15.04\textbf{(64\%$\uparrow$)} & 11.23\textbf{(64\%$\uparrow$)} \\
		\hline
			OURS  &  33.27  & 24.80 &  24.70  & 18.47 \\
		\hline
	\end{tabular}%
	}
	
	 \subfloat[The visualization version of results reported in the above Table. From this figure we can intuitively observe that improvements achieved by our framework of complex subsets~(MH-SO, SH-MO, MH-MO) are higher than the simple one~(SH-SO).
	\label{fig:subset-fig}]{
		\centering

    \begin{tabular}{c }
   
     \includegraphics[width=320pt]{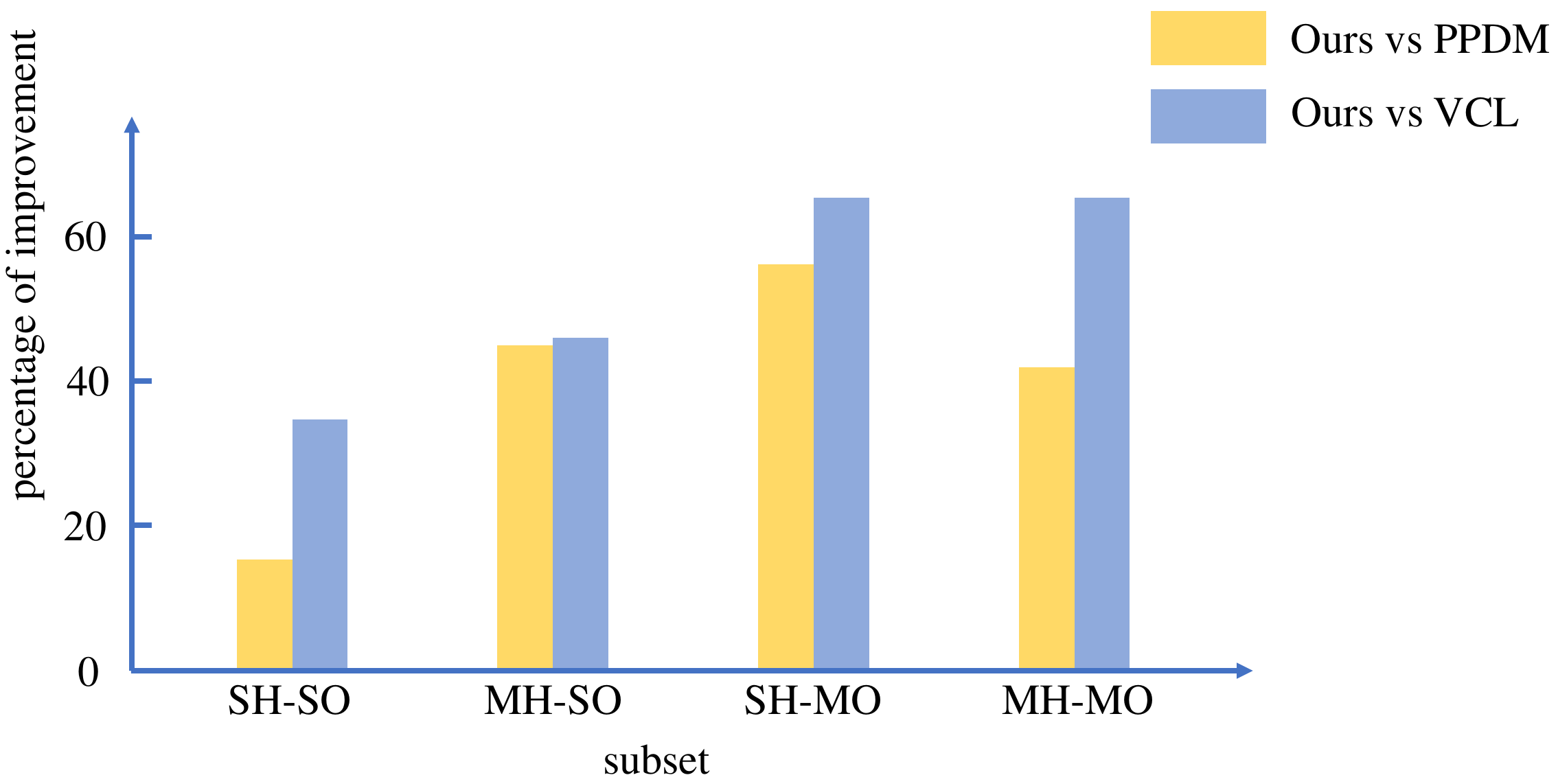}  \\
     
    \end{tabular}

	}
		
	\caption{Results on different subsets. These subsets are generated according to the number of annotated persons and objects in the image, \textit{i.e.}, single-human \& single-object(SH-SO), multi-human \& single-object(MH-SO), single-human \& multi-object(SH-MO) and multi-human \& multi-object(MH-MO). Our method obtains consistent improvement on all subsets and is especially more robust in complex scenes.
	}
	\label{tab:simple-complex set result}
	\vspace{1mm}
\end{table*}

\begin{figure}[!htb]
	\centering
	\includegraphics[width=\linewidth]{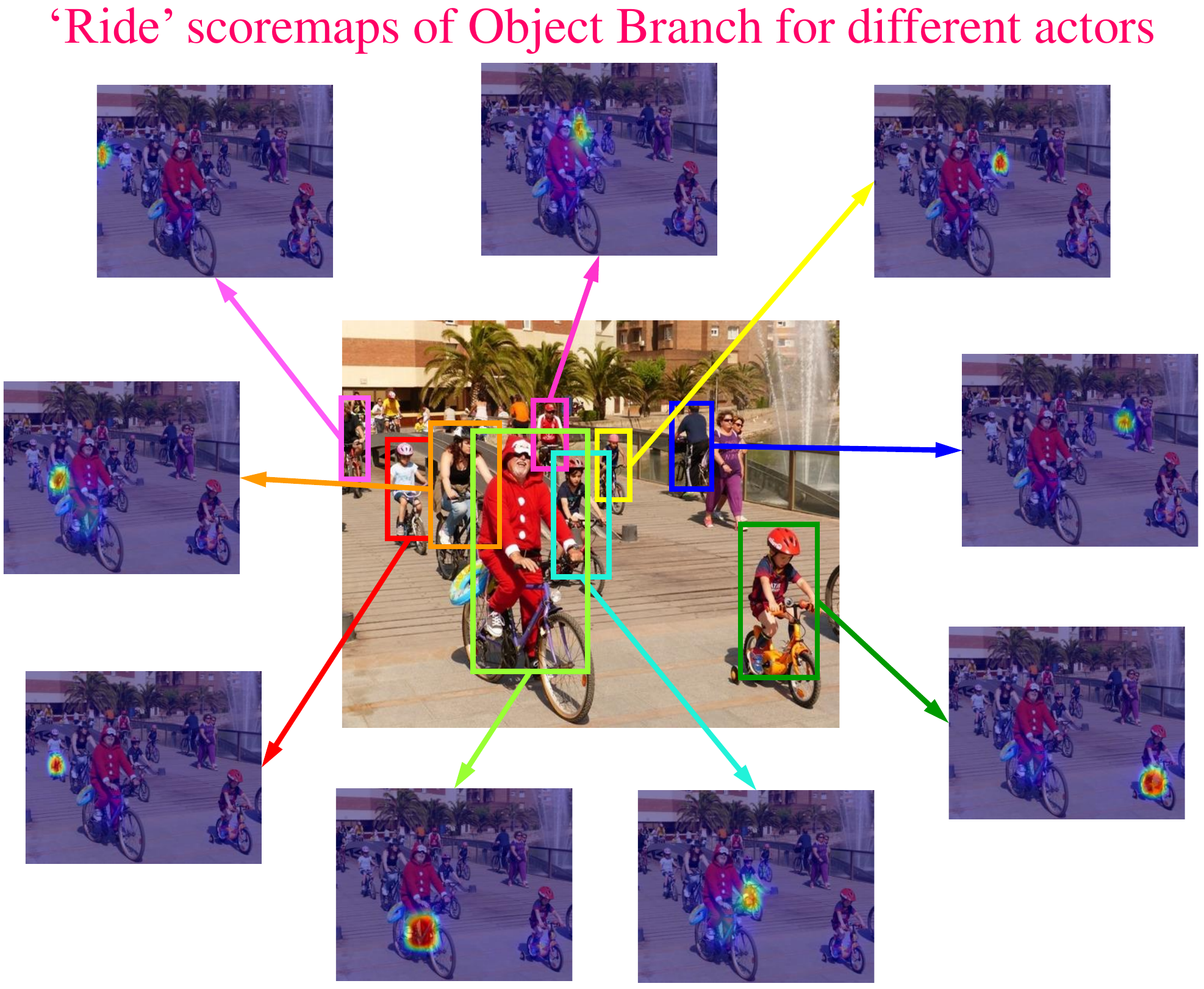}
	\caption{The heatmap results of Object Branch for different actors in the same image. Boxes with different colors mean different detected persons. When different individuals serve as the actor, the resulting heatmaps are shown where the colored arrows point. It has been observed that our model has the ability to localize the corresponding interacting regions for each actor accurately in complex scenes with multiple interactions.}
% 	\vspace{0.2cm}
	\label{fig:scoremap}
\end{figure}

\subsection{Analysis and Discussion}
\noindent
\textbf{Results on simple and complex subsets}
To evaluate the robustness of our proposed idea for data with different complexity, we split the HICO-DET into four subsets according to the number of annotated persons and objects, \textit{i.e.}, multi-human \& multi-object(MH-MO), single-human \& multi-object(SH-MO), multi-human \& single-object(MH-SO) and single-human \& single-object(SH-SO). We compare our proposed framework with PPDM and VCL on these four subsets in Tab.~\ref{tab:simple-complex set result}. Results intuitively show that our method consistently obtains better performance on all four subsets. Especially for complex data~(SH-MO, MH-SO, MH-MO), our method achieves more improvement. That is, our method is consistent effective and more suitable for complex scenes.

\noindent
\textbf{Qualitative Result}
Some qualitative results compared with PPDM and VCL in shown in Fig. \ref{fig:visual}. 
% PPDM is a state-of-the-art One-Stage HOI detection method and VCL is a standard Crop-Based method. 
We select several images in the HICO-DET and V-COCO datasets to show the ability of our proposed framework of addressing the multiple interaction occurrence issues. These cases are quite complex and easy to cause ambiguities for crop-based method, and highly-coupled approach. As illustrated in Fig. \ref{fig:visual}, our approach handles the complex crowded scenes, which are failed by the two other approaches. And the Object Scoremap shown in the last row further validates the effectiveness of our model to capture the interactions between actor and objects. 

Moreover, we also illustrate the heatmap results of the Object Branch for different actors in the same image as shown in Fig. \ref{fig:scoremap}. The high consistency between actors and object regions demonstrates that our approach has a powerful ability to disambiguate multiple complex interactions in the same image.

\section{Conclusion}
In this paper, we have developed an effective actor-centric approach for human-object interaction detection. Our approach formulates the task as a pixel-wise prediction problem, and learns non-local features by introducing binary masks of persons. The relationship between one person and multiple objects of the entire image has been explored to promote the interaction reasoning in a unified framework. 
For enhancing the learning effect under the proposed framework, we further introduce a Weighted Cross Entropy Loss consisting of Hanning weight and Scale weight.
Experimental results have shown that our approach achieves state-of-the-art performance in both HICO-DET and V-COCO datasets, especially on complex images with multiple persons/objects.

\section{Declaration of Competing Interest}
The authors declare that they have no known competing financial interests or personal relationships that could have appeared to influence the work reported in this paper.
\section{Acknowledgements}
This work is supported by the National Natural Science Foundation of China (NSFC) grant 62176100, the Special Project of Science and Technology Development of Central guiding Local grant 2021BEE056.

\bibliography{mybibfile}

\begin{thebibliography}{10}
\expandafter\ifx\csname url\endcsname\relax
  \def\url#1{\texttt{#1}}\fi
\expandafter\ifx\csname urlprefix\endcsname\relax\def\urlprefix{URL }\fi
\expandafter\ifx\csname href\endcsname\relax
  \def\href#1#2{#2} \def\path#1{#1}\fi

\bibitem{chen2014predicting}
C.-Y. Chen, K.~Grauman, Predicting the location of “interactees” in novel
  human-object interactions, in: Asian conference on computer vision, Springer,
  2014, pp. 351--367.

\bibitem{yao2010modeling}
B.~Yao, L.~Fei-Fei, Modeling mutual context of object and human pose in
  human-object interaction activities, in: 2010 IEEE Computer Society
  Conference on Computer Vision and Pattern Recognition, IEEE, 2010, pp.
  17--24.

\bibitem{gao2020interactgan}
C.~Gao, S.~Liu, D.~Zhu, Q.~Liu, J.~Cao, H.~He, R.~He, S.~Yan, Interactgan:
  Learning to generate human-object interaction, in: Proceedings of the 28th
  ACM International Conference on Multimedia, 2020, pp. 165--173.

\bibitem{aksoy2011learning}
E.~E. Aksoy, A.~Abramov, J.~D{\"o}rr, K.~Ning, B.~Dellen, F.~W{\"o}rg{\"o}tter,
  Learning the semantics of object--action relations by observation, The
  International Journal of Robotics Research 30~(10) (2011) 1229--1249.

\bibitem{worgotter2013simple}
F.~W{\"o}rg{\"o}tter, E.~E. Aksoy, N.~Kr{\"u}ger, J.~Piater, A.~Ude,
  M.~Tamosiunaite, A simple ontology of manipulation actions based on
  hand-object relations, IEEE Transactions on Autonomous Mental Development
  5~(2) (2013) 117--134.

\bibitem{argall2009survey}
B.~D. Argall, S.~Chernova, M.~Veloso, B.~Browning, A survey of robot learning
  from demonstration, Robotics and autonomous systems 57~(5) (2009) 469--483.

\bibitem{yang2013detection}
Y.~Yang, C.~Fermuller, Y.~Aloimonos, Detection of manipulation action
  consequences (mac), in: Proceedings of the IEEE Conference on Computer Vision
  and Pattern Recognition, 2013, pp. 2563--2570.

\bibitem{gkioxari2018detecting}
G.~Gkioxari, R.~Girshick, P.~Doll{\'a}r, K.~He, Detecting and recognizing
  human-object interactions, in: Proceedings of the IEEE Conference on Computer
  Vision and Pattern Recognition, 2018, pp. 8359--8367.

\bibitem{hou2020visual}
Z.~Hou, X.~Peng, Y.~Qiao, D.~Tao, Visual compositional learning for
  human-object interaction detection, in: European Conference on Computer
  Vision, Springer, 2020, pp. 584--600.

\bibitem{chao2018learning}
Y.-W. Chao, Y.~Liu, X.~Liu, H.~Zeng, J.~Deng, Learning to detect human-object
  interactions, in: 2018 ieee winter conference on applications of computer
  vision (wacv), IEEE, 2018, pp. 381--389.

\bibitem{xu2019learning}
B.~Xu, Y.~Wong, J.~Li, Q.~Zhao, M.~S. Kankanhalli, Learning to detect
  human-object interactions with knowledge, in: Proceedings of the IEEE/CVF
  Conference on Computer Vision and Pattern Recognition, 2019.

\bibitem{liu2020consnet}
Y.~Liu, J.~Yuan, C.~W. Chen, Consnet: Learning consistency graph for zero-shot
  human-object interaction detection, in: Proceedings of the 28th ACM
  International Conference on Multimedia, 2020, pp. 4235--4243.

\bibitem{gao2018ican}
C.~Gao, Y.~Zou, J.-B. Huang, ican: Instance-centric attention network for
  human-object interaction detection, arXiv preprint arXiv:1808.10437.

\bibitem{ulutan2020vsgnet}
O.~Ulutan, A.~Iftekhar, B.~S. Manjunath, Vsgnet: Spatial attention network for
  detecting human object interactions using graph convolutions, in: Proceedings
  of the IEEE/CVF Conference on Computer Vision and Pattern Recognition, 2020,
  pp. 13617--13626.

\bibitem{liao2020ppdm}
Y.~Liao, S.~Liu, F.~Wang, Y.~Chen, C.~Qian, J.~Feng, Ppdm: Parallel point
  detection and matching for real-time human-object interaction detection, in:
  Proceedings of the IEEE/CVF Conference on Computer Vision and Pattern
  Recognition, 2020, pp. 482--490.

\bibitem{wang2019deep}
T.~Wang, R.~M. Anwer, M.~H. Khan, F.~S. Khan, Y.~Pang, L.~Shao, J.~Laaksonen,
  Deep contextual attention for human-object interaction detection, in:
  Proceedings of the IEEE/CVF International Conference on Computer Vision,
  2019, pp. 5694--5702.

\bibitem{wang2020learning}
T.~Wang, T.~Yang, M.~Danelljan, F.~S. Khan, X.~Zhang, J.~Sun, Learning
  human-object interaction detection using interaction points, in: Proceedings
  of the IEEE/CVF Conference on Computer Vision and Pattern Recognition, 2020,
  pp. 4116--4125.

\bibitem{zhong2021glance}
X.~Zhong, X.~Qu, C.~Ding, D.~Tao, Glance and gaze: Inferring action-aware
  points for one-stage human-object interaction detection, in: Proceedings of
  the IEEE/CVF Conference on Computer Vision and Pattern Recognition, 2021, pp.
  13234--13243.

\bibitem{zhang2021mining}
A.~Zhang, Y.~Liao, S.~Liu, M.~Lu, Y.~Wang, C.~Gao, X.~Li, Mining the benefits
  of two-stage and one-stage hoi detection, in: Thirty-Fifth Conference on
  Neural Information Processing Systems, 2021.

\bibitem{ASNET}
M.~Chen, Y.~Liao, S.~Liu, Z.~Chen, F.~Wang, C.~Qian, Reformulating hoi
  detection as adaptive set prediction, in: Proceedings of the IEEE/CVF
  Conference on Computer Vision and Pattern Recognition, 2021, pp. 9004--9013.

\bibitem{li2021improving}
Z.~Li, C.~Zou, Y.~Zhao, B.~Li, S.~Zhong, Improving human-object interaction
  detection via phrase learning and label composition, arXiv preprint
  arXiv:2112.07383.

\bibitem{newell2016stacked}
A.~Newell, K.~Yang, J.~Deng, Stacked hourglass networks for human pose
  estimation, in: European conference on computer vision, Springer, 2016, pp.
  483--499.

\bibitem{law2018cornernet}
H.~Law, J.~Deng, Cornernet: Detecting objects as paired keypoints, in:
  Proceedings of the European conference on computer vision (ECCV), 2018, pp.
  734--750.

\bibitem{lin2017feature}
T.-Y. Lin, P.~Doll{\'a}r, R.~Girshick, K.~He, B.~Hariharan, S.~Belongie,
  Feature pyramid networks for object detection, in: Proceedings of the IEEE
  conference on computer vision and pattern recognition, 2017, pp. 2117--2125.

\bibitem{yu2018deep}
F.~Yu, D.~Wang, E.~Shelhamer, T.~Darrell, Deep layer aggregation, in:
  Proceedings of the IEEE conference on computer vision and pattern
  recognition, 2018, pp. 2403--2412.

\bibitem{sun2019deep}
K.~Sun, B.~Xiao, D.~Liu, J.~Wang, Deep high-resolution representation learning
  for human pose estimation, in: Proceedings of the IEEE/CVF Conference on
  Computer Vision and Pattern Recognition, 2019, pp. 5693--5703.

\bibitem{gupta2015visual}
S.~Gupta, J.~Malik, Visual semantic role labeling, arXiv preprint
  arXiv:1505.04474.

\bibitem{lin2014microsoft}
T.-Y. Lin, M.~Maire, S.~Belongie, J.~Hays, P.~Perona, D.~Ramanan,
  P.~Doll{\'a}r, C.~L. Zitnick, Microsoft coco: Common objects in context, in:
  European conference on computer vision, Springer, 2014, pp. 740--755.

\bibitem{qi2018learning}
S.~Qi, W.~Wang, B.~Jia, J.~Shen, S.-C. Zhu, Learning human-object interactions
  by graph parsing neural networks, in: Proceedings of the European Conference
  on Computer Vision (ECCV), 2018, pp. 401--417.

\bibitem{zhou2019relation}
P.~Zhou, M.~Chi, Relation parsing neural network for human-object interaction
  detection, in: Proceedings of the IEEE/CVF International Conference on
  Computer Vision, 2019, pp. 843--851.

\bibitem{li2019tin}
Y.-L. Li, S.~Zhou, X.~Huang, L.~Xu, Z.~Ma, H.-S. Fang, Y.~Wang, C.~Lu,
  Transferable interactiveness knowledge for human-object interaction
  detection, in: Proceedings of the IEEE Conference on Computer Vision and
  Pattern Recognition, 2019, pp. 3585--3594.

\bibitem{zhou2020cascaded}
T.~Zhou, W.~Wang, S.~Qi, H.~Ling, J.~Shen, Cascaded human-object interaction
  recognition, in: Proceedings of the IEEE/CVF Conference on Computer Vision
  and Pattern Recognition, 2020, pp. 4263--4272.

\bibitem{li2019transferable}
Y.-L. Li, S.~Zhou, X.~Huang, L.~Xu, Z.~Ma, H.-S. Fang, Y.~Wang, C.~Lu,
  Transferable interactiveness knowledge for human-object interaction
  detection, in: Proceedings of the IEEE/CVF Conference on Computer Vision and
  Pattern Recognition, 2019, pp. 3585--3594.

\bibitem{li2020pastanet}
Y.-L. Li, L.~Xu, X.~Liu, X.~Huang, Y.~Xu, S.~Wang, H.-S. Fang, Z.~Ma, M.~Chen,
  C.~Lu, Pastanet: Toward human activity knowledge engine, in: Proceedings of
  the IEEE/CVF Conference on Computer Vision and Pattern Recognition, 2020, pp.
  382--391.

\bibitem{gao2020drg}
C.~Gao, J.~Xu, Y.~Zou, J.-B. Huang, Drg: Dual relation graph for human-object
  interaction detection, in: European Conference on Computer Vision, Springer,
  2020, pp. 696--712.

\bibitem{feng2019turbo}
W.~Feng, W.~Liu, T.~Li, J.~Peng, C.~Qian, X.~Hu, Turbo learning framework for
  human-object interactions recognition and human pose estimation, in:
  Proceedings of the AAAI Conference on Artificial Intelligence, 2019, pp.
  898--905.

\bibitem{eccv2020actioncoprior}
D.-J. Kim, X.~Sun, J.~Choi, S.~Lin, I.~S. Kweon, Detecting human-object
  interactions with action co-occurrence priors, arXiv preprint
  arXiv:2007.08728.

\bibitem{li2020hoi}
Y.-L. Li, X.~Liu, X.~Wu, Y.~Li, C.~Lu, Hoi analysis: Integrating and
  decomposing human-object interaction, Advances in Neural Information
  Processing Systems 33.

\bibitem{ren2016faster}
S.~Ren, K.~He, R.~Girshick, J.~Sun, Faster r-cnn: towards real-time object
  detection with region proposal networks, IEEE transactions on pattern
  analysis and machine intelligence 39~(6) (2016) 1137--1149.

\bibitem{chen2019mmdetection}
K.~Chen, J.~Wang, J.~Pang, Y.~Cao, Y.~Xiong, X.~Li, S.~Sun, W.~Feng, Z.~Liu,
  J.~Xu, et~al., Mmdetection: Open mmlab detection toolbox and benchmark, arXiv
  preprint arXiv:1906.07155.

\bibitem{wan2019pose}
B.~Wan, D.~Zhou, Y.~Liu, R.~Li, X.~He, Pose-aware multi-level feature network
  for human object interaction detection, in: Proceedings of the IEEE/CVF
  International Conference on Computer Vision, 2019, pp. 9469--9478.

\bibitem{gupta2019no}
T.~Gupta, A.~Schwing, D.~Hoiem, No-frills human-object interaction detection:
  Factorization, layout encodings, and training techniques, in: Proceedings of
  the IEEE/CVF International Conference on Computer Vision, 2019, pp.
  9677--9685.

\bibitem{eccv2020hetegraph}
H.~Wang, W.-s. Zheng, L.~Yingbiao, Contextual heterogeneous graph network for
  human-object interaction detection, arXiv preprint arXiv:2010.10001.

\bibitem{kim2020uniondet}
B.~Kim, T.~Choi, J.~Kang, H.~J. Kim, Uniondet: Union-level detector towards
  real-time human-object interaction detection, in: European Conference on
  Computer Vision, Springer, 2020, pp. 498--514.

\bibitem{peyre2019detecting}
J.~Peyre, I.~Laptev, C.~Schmid, J.~Sivic, Detecting unseen visual relations
  using analogies, in: Proceedings of the IEEE International Conference on
  Computer Vision, 2019, pp. 1981--1990.

\bibitem{eccv2020keycues}
Y.~Liu, Q.~Chen, A.~Zisserman, Amplifying key cues for human-object-interaction
  detection, in: European Conference on Computer Vision, Springer, 2020, pp.
  248--265.

\bibitem{bansal2020detecting}
A.~Bansal, S.~S. Rambhatla, A.~Shrivastava, R.~Chellappa, Detecting
  human-object interactions via functional generalization., in: AAAI, 2020, pp.
  10460--10469.

\bibitem{kim2021hotr}
B.~Kim, J.~Lee, J.~Kang, E.-S. Kim, H.~J. Kim, Hotr: End-to-end human-object
  interaction detection with transformers, in: Proceedings of the IEEE/CVF
  Conference on Computer Vision and Pattern Recognition, 2021, pp. 74--83.

\bibitem{zou2021end}
C.~Zou, B.~Wang, Y.~Hu, J.~Liu, Q.~Wu, Y.~Zhao, B.~Li, C.~Zhang, C.~Zhang,
  Y.~Wei, et~al., End-to-end human object interaction detection with hoi
  transformer, in: Proceedings of the IEEE/CVF Conference on Computer Vision
  and Pattern Recognition, 2021, pp. 11825--11834.

\bibitem{russakovsky2015imagenet}
O.~Russakovsky, J.~Deng, H.~Su, J.~Krause, S.~Satheesh, S.~Ma, Z.~Huang,
  A.~Karpathy, A.~Khosla, M.~Bernstein, et~al., Imagenet large scale visual
  recognition challenge, International journal of computer vision 115~(3)
  (2015) 211--252.

\bibitem{kingma2014adam}
D.~P. Kingma, J.~Ba, Adam: A method for stochastic optimization, arXiv preprint
  arXiv:1412.6980.

\end{thebibliography}

\end{document}